\documentclass[10pt,twocolumn,letterpaper]{article}

\usepackage[pagenumbers]{cvpr} % To force page numbers, e.g. for an arXiv version

\usepackage{float}
\usepackage{caption}
\usepackage{multirow}
\usepackage{dblfloatfix}
\definecolor{cvprblue}{rgb}{0.21,0.49,0.74}
\usepackage[pagebackref,breaklinks,colorlinks,allcolors=cvprblue]{hyperref}

\def\confName{CVPR}
\def\confYear{2027}

\title{PRISM: Feed-Forward Single-Image 3D Reconstruction via Geometric Warp-Residual Modeling}

\author{Zhijie Zheng, Xinhao Xiang, Jiawei Zhang\\
University of California, Davis\\
CA 95616, USA\\
{\tt\small \{zhjzheng, xhxiang, jiwzhang\}@ucdavis.edu}
}

\begin{document}

% \maketitle
\twocolumn[{
  \maketitle
  \begin{center}
    \includegraphics[width=1.0\textwidth]{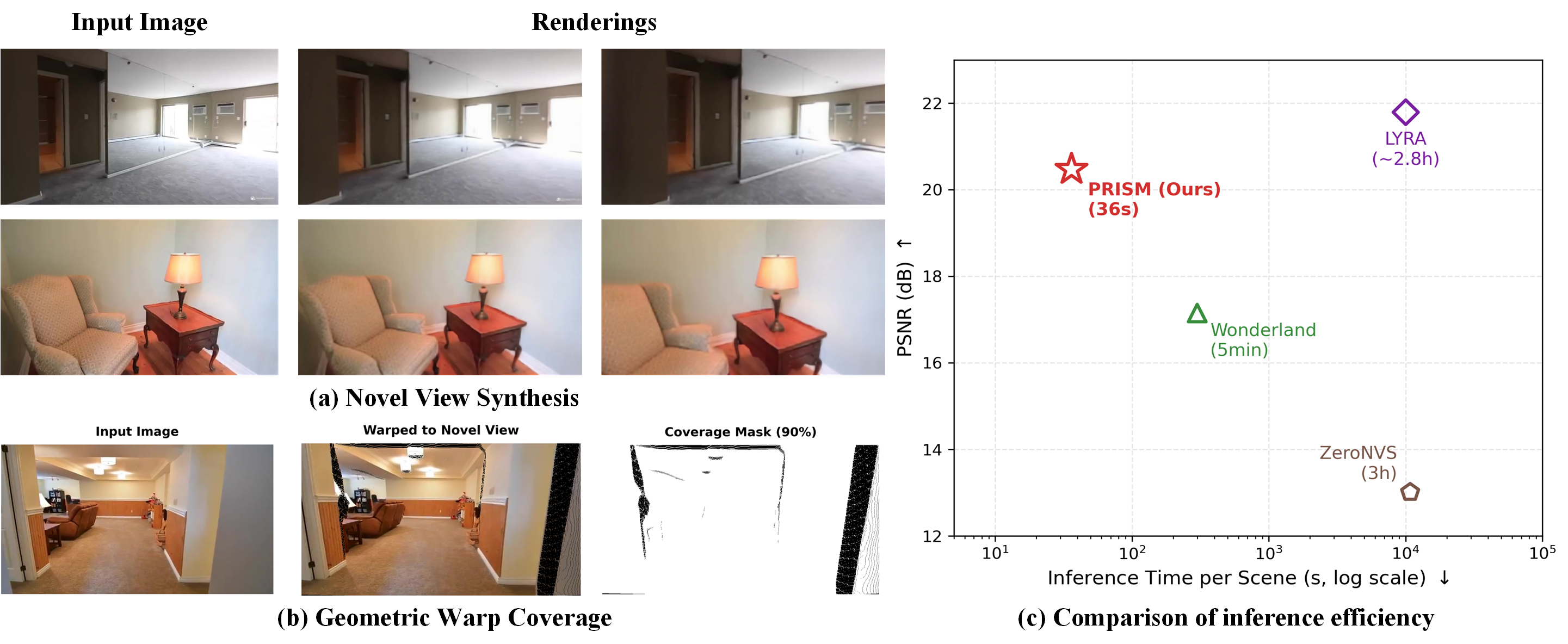}
    \captionof{figure}{\textbf{Single-image 3D reconstruction and efficiency comparison.}
(a) Given a single input image, PRISM synthesizes geometrically consistent novel views across diverse scenes.
(b) Geometric forward warping covers most of a target view directly from the input, with only a small residual left for the encoder to correct.
(c) PRISM achieves competitive reconstruction quality on RealEstate10K~\cite{zhou2018stereo} at $277\times$ lower inference time cost than LYRA~\cite{bahmani2025lyra}, taking just 36 seconds per scene.}
  \label{fig:illustration}
  \end{center}
}]

\begin{abstract}
Reconstructing 3D scenes from a single image is a fundamental challenge in computer vision, with broad applications in virtual reality, robotics, and content creation. Recent methods achieve outstanding performance by leveraging camera-controlled video diffusion models, but rely on iterative diffusion sampling, which greatly limits their practical deployment. We observe that geometric forward warping alone can cover the majority of a target view directly from the input image, with only a compact residual left for the encoder to correct. Motivated by this observation, we propose \textbf{PRISM}, a feed-forward framework that decomposes multi-view latent prediction into a parameter-free geometric prior and a learned residual correction, with no diffusion sampling required at inference. To enable generalization from purely synthetic training data, we devise a two-stage training strategy combining latents supervised distillation for geometric generalization and perceptual fine-tuning for appearance quality optimization. Extensive experiments on three benchmarks demonstrate that PRISM achieves competitive reconstruction quality compared with diffusion-based methods, while reducing inference time dramatically to only 36 seconds per scene.
Our code is available at \url{https://github.com/microjie372/PRISM}.
\end{abstract}    
\section{Introduction}
\label{sec:intro}

\begin{figure*}[htbp]
	\centering
	\includegraphics[width=0.8\linewidth]{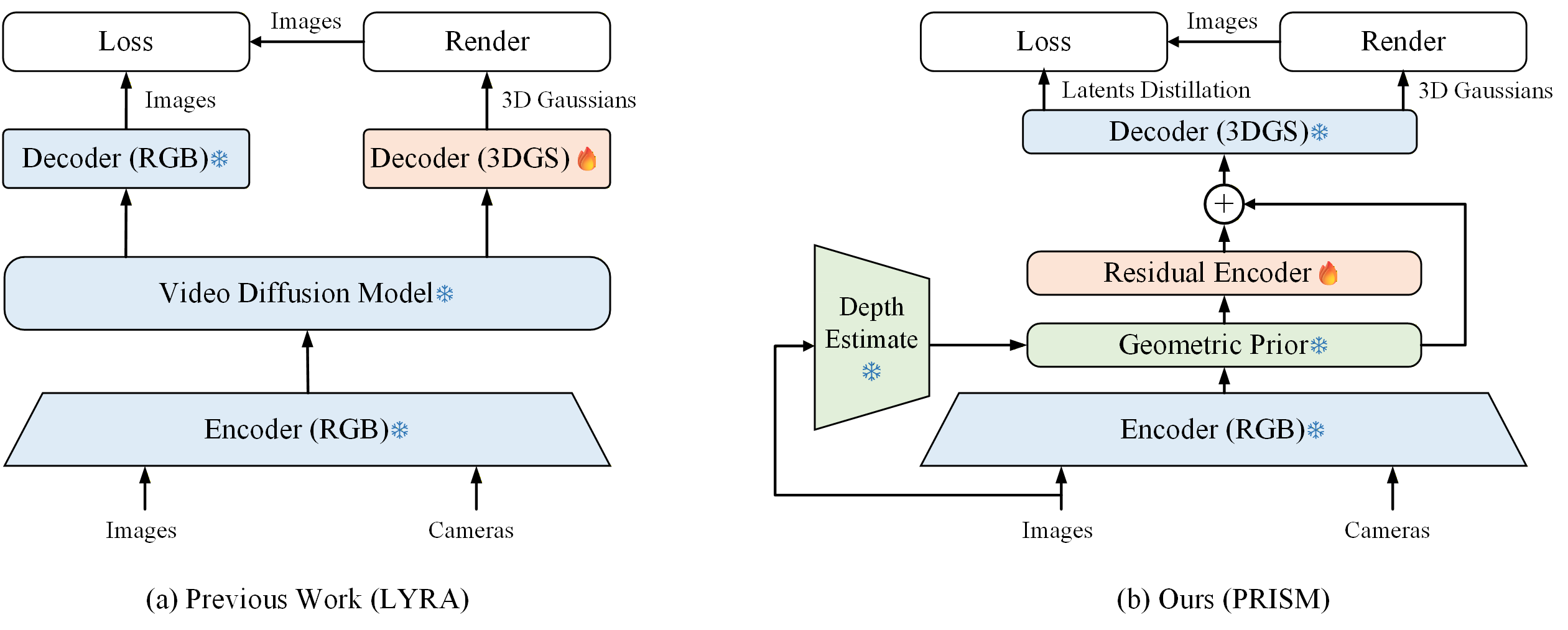}
	\caption{\textbf{From diffusion sampling to feed-forward geometric reasoning.} LYRA~\cite{bahmani2025lyra} (a) relies on a frozen video diffusion model to generate multi-view latents at inference, decoded into 3D Gaussians by a trainable 3DGS decoder. In contrast, PRISM (b) replaces the diffusion model with a lightweight residual encoder that predicts residual corrections to a parameter-free geometric prior derived from monocular depth, trained via latents supervised distillation from the frozen 3DGS decoder.}
	\label{fig:structure}
    \vspace{-0.1cm}
\end{figure*}

High-quality 3D scene reconstruction has become a critical capability in computer vision and graphics, with broad applications in physical AI, robotic simulation, and immersive content creation.
These applications demand explicit 3D representations that support real-time rendering, geometric consistency, and scalable generation across diverse scenes.
Neural rendering methods such as NeRF~\cite{mildenhall2020nerf, muller2022instant} and 3DGS~\cite{kerbl20233dgs, lu2024scaffoldgs} recover high-fidelity representations from posed images through per-scene optimization, but require dense multi-view captures and lengthy optimization times, which significantly limits their scalability.
Recent feed-forward reconstruction models~\cite{wang2024dust3r, wang2025vggt, hong2024lrm, zhang2024gslrm, leroy2024mast3r}, such as VGGT~\cite{wang2025vggt} and DA3~\cite{lin2025da3}, amortize reconstruction across scenes and eliminate per-scene optimization, yet still require multiple images as input.
The ability to reconstruct a complete 3D scene from a single uncalibrated image, without additional views or camera parameters, offers a more accessible path to real-world 3D understanding.
% and scalable

Video diffusion models~\cite{cosmos2025, wan2025} trained on internet-scale video data implicitly encode rich geometric knowledge of the 3D world.
By learning from real-world videos from indoor scenes to large-scale outdoor environments, these models achieve impressive generalization and can hallucinate plausible content well beyond what is visible in the input frame.
Camera control mechanisms~\cite{ren2025gen3c} further enable explicit conditioning on target viewpoints, and transform video diffusion models into powerful tools for novel view synthesis.
Recent methods exploit this prior for single-image 3D scene reconstruction: LYRA~\cite{bahmani2025lyra}, Wonderland~\cite{liang2025wonderland}, and Bolt3D~\cite{szymanowicz2025bolt3d} decode video diffusion latents into 3D Gaussian primitives from a single image.
However, iterative diffusion sampling across hundreds of video frames may take hours per scene at inference~\cite{bahmani2025lyra}, making these methods impractical for large-scale deployment.

% We observe that diffusion is not the only means to populate a target view.
We observe that much of a target view can also be populated through geometric forward warping.
As shown in Fig.~\ref{fig:illustration}(b), given monocular depth estimation~\cite{wang2025moge}, geometric forward warping can already cover up to 90\% of a target view directly from the input image, measured on RE10K scenes in which the target viewpoint stays close to the reference.
The remaining residual, arising from disoccluded regions and depth estimation errors, constitutes a small fraction of the total latent on average, which leaves a tractable correction for a lightweight feed-forward encoder.
This geometric decomposition fundamentally changes the nature of the prediction problem. Instead of generating all target view content from scratch via diffusion, the encoder need only predict a compact correction over a strong geometric prior. Warp coverage decreases as the baseline between the reference and target views widens, which in turn enlarges the residual the encoder must predict.

Motivated by this observation, we propose \textbf{PRISM}, which replaces the video diffusion model at inference with a lightweight residual encoder, as illustrated in Fig.~\ref{fig:structure}.
PRISM decomposes multi-view latent prediction into a parameter-free geometric prior, obtained directly from depth-based forward warping in the video latent space, and a learned residual correction predicted by a compact encoder.
Moreover, PRISM operates in a compressed video latent space rather than pixel space, matching the representation its supervision and frozen decoder already use.
The predicted latents are decoded by a frozen 3DGS decoder into a complete 3D scene, with no diffusion sampling at inference.
To enable generalization from purely synthetic training data to real-world scenes, we devise a two-stage training strategy.
In the first stage, latents supervised distillation from the frozen reconstruction decoder aligns the encoder's predictions with the decoder's expected latent distribution. In the second stage, perceptual fine-tuning directly optimizes rendered appearance quality. As shown in Fig.~\ref{fig:illustration}(c), this design places PRISM at a different point on the quality-efficiency curve compared to existing methods.

Our main contributions are as follows:
\begin{itemize}
    \item We propose PRISM, which decomposes single-image 3D reconstruction into a parameter-free geometric prior and a learned residual correction, eliminating diffusion sampling at inference.
    \item We devise a two-stage training strategy combining latents supervised distillation and perceptual fine-tuning, so that PRISM generalizes from synthetic training data to real-world scenes without any real multi-view supervision.
    \item Extensive experiments on three benchmarks demonstrate that PRISM achieves reconstruction quality competitive with diffusion-based methods, with inference time reduced to 36 seconds per scene, a $277\times$ speedup compared to LYRA.
\end{itemize}
\section{Related Work}
\label{sec:relatedwork}

\paragraph{Single-Image 3D Reconstruction.}
Early methods optimize NeRF~\cite{mildenhall2020nerf} or 3DGS~\cite{kerbl20233dgs} per scene, which limits scalability.
Feed-forward models such as GS-LRM~\cite{zhang2024gslrm}, DUSt3R~\cite{wang2024dust3r}, MASt3R~\cite{leroy2024mast3r}, VGGT~\cite{wang2025vggt} and AVGGT~\cite{sun2026avggt} eliminate per-scene optimization but are designed for multi-view input. Among methods that minimize input requirements, LRM~\cite{hong2024lrm} and Splatter Image~\cite{szymanowicz2024splatterimage} achieve object-level reconstruction from a single view, while scene-level methods~\cite{charatan2024pixelsplat, chen2024mvsplat, xu2025depthsplat, jiang2025anysplat, itkin2026globalsplat, li2026tokensplat} still require at least two views. Flash3D~\cite{szymanowicz20253dv} extends monocular depth estimation to scene-level single-image Gaussian prediction, but operates in pixel space without leveraging a pretrained reconstruction backbone. Diffusion-based approaches exploit generative priors for single-image reconstruction~\cite{liu2023zero123, shi2023zero123plus, liu2024syncdreamer, liu2023one2345, long2024wonder3d, wu2024reconfusion, gao2024cat3d, sargent2024zeronvs}, and recent scene-level methods extend this paradigm further: ViewCrafter~\cite{yu2025viewcrafter} adds per-scene optimization, Wonderland~\cite{liang2025wonderland} and LYRA~\cite{bahmani2025lyra} decode video diffusion latents into 3DGS through feed-forward networks, and Bolt3D~\cite{szymanowicz2025bolt3d} amortizes reconstruction via a purpose-built multi-view latent diffusion model.
Nevertheless, all of these methods require diffusion sampling at inference time.
PRISM eliminates diffusion sampling at inference through deterministic geometric reasoning, and achieves competitive reconstruction quality at much lower inference cost.

\paragraph{Geometry-Guided View Synthesis.}
Depth-based forward warping provides a parameter-free geometric prior for novel view synthesis that enables classical methods~\cite{tucker2020mpi, shih20203d, wiles2020synsin, cao2022fwd, rockwell2021pixelsynth} to relocate reference pixels to target viewpoints using estimated depth.
Despite their simplicity, these methods perform warping in pixel space, where sub-pixel depth errors produce visible artifacts, and require a separate inpainting stage to fill disoccluded regions.
Learning-based approaches such as pixelNeRF~\cite{yu2021pixelnerf}, IBRNet~\cite{wang2021ibrnet}, and MVSNeRF~\cite{chen2021mvsnerf} instead learn to aggregate multi-view features for novel view synthesis, but require multiple posed images at inference and do not exploit geometric warping.
Building on this, GenWarp~\cite{seo2024genwarp} integrates depth-based warping with a text-to-image diffusion model and trains it to jointly decide where to warp and where to generate, which removes the need for explicit hole-filling.
% #---# R4: warp-based NVS is well studied; distinguish from earlier work.
% #---# WAVE cite added, checked against arXiv 2506.23518.
WAVE~\cite{park2025wave} guides a pretrained diffusion sampler with warped images through training-free attention manipulation and noise reinitialization, improving view consistency.
Nevertheless, the reliance on pixel-space warping and a generative completion stage persists.
In contrast, PRISM performs warping directly in the compressed video latent space where the frozen decoder operates, so the prior needs no pixel-space inpainting stage.
A single encoder then jointly corrects geometric misalignment and fills missing regions in a single forward pass.

\paragraph{Distillation and Acceleration of Generative Models.}
Diffusion probabilistic models~\cite{ho2020ddpm} generate high-quality outputs through iterative denoising, while latent diffusion models~\cite{rombach2022ldm} reduce cost by working in compressed latent spaces.
Substantial effort has been devoted to reducing sampling steps, including deterministic ODE solvers~\cite{song2021ddim, lu2022dpmsolver}, progressive distillation~\cite{salimans2022progressive}, consistency models~\cite{song2023consistency}, and flow-matching-based approaches~\cite{lipman2023flow, liu2023rectifiedflow} that linearize the sampling trajectory.
In the 3D domain, VideoScene~\cite{wang2025videoscene} applies flow distillation to reduce scene-level video diffusion to a single inference step, and VFusion3D~\cite{han2024vfusion3d} demonstrate that diffusion outputs can supervise a feed-forward 3D model offline, decoupling inference from the generative pipeline.
Unlike VFusion3D, which supervises a feed-forward model directly on diffusion-generated multi-view images for object-level reconstruction, PRISM exploits a geometric warp prior to decompose scene-level latent prediction into a parameter-free prior and a compact residual.

% #---# R1: Fig 3 caption should name the Cosmos encoder.
\begin{figure*}[tbp]
    \centering
    \includegraphics[width=\linewidth]{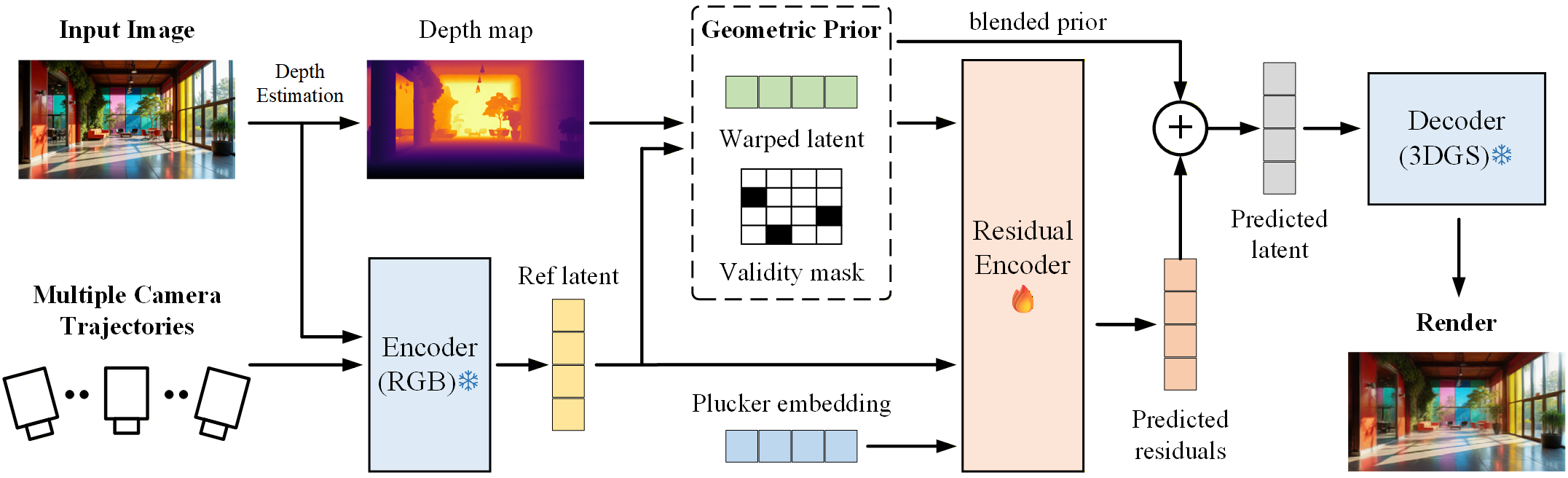}
    % \caption{\textbf{Overview of PRISM.} Given a single input image and multiple camera trajectories, PRISM estimates monocular depth and encodes the reference image into a latent representation. The depth map guides z-buffered forward splatting in the Cosmos latent space to form the geometric prior. The Residual Encoder takes the warped latent, validity mask, reference latent, and Pl\"{u}cker embeddings as input and predicts a per-view residual correction. The predicted residuals are added to the blended prior to obtain the predicted latents, which are decoded by the frozen 3DGS decoder into a 3D scene for novel view rendering.}
    \caption{\textbf{Overview of PRISM.} Given a single input image and multiple camera trajectories, PRISM estimates monocular depth and encodes the reference image into a latent representation using the frozen Cosmos VAE encoder. The depth map guides z-buffered forward splatting in the Cosmos latent space to form the geometric prior. The Residual Encoder takes the warped latent, validity mask, reference latent, and Pl\"{u}cker embeddings as input and predicts a per-view residual correction. The predicted residuals are added to the blended prior to obtain the predicted latents, which are decoded by the frozen 3DGS decoder into a 3D scene for novel view rendering.}
    \label{fig:framework}
    \vspace{-0.1cm}
\end{figure*}

\section{Method}
\label{sec:method}

\subsection{Preliminaries}
\label{sec:prelim}

\noindent\textbf{Task definition.}
Given a single RGB image $\mathbf{I} \in \mathbb{R}^{H \times W \times 3}$ and $V$ target camera trajectories $\{\boldsymbol{\pi}^v\}_{v=1}^{V}$, where each trajectory $\boldsymbol{\pi}^v = \{c^v_t\}_{t=1}^{L}$ specifies $L$ camera poses, our goal is to synthesize novel views $\{\hat{\mathbf{x}}^v_t\}$ at all target viewpoints and timesteps.
Following LYRA~\cite{bahmani2025lyra}, we represent the scene as 3D Gaussian primitives $\mathcal{G}$, which support efficient differentiable rendering at arbitrary camera poses.
The Pl\"{u}cker ray embeddings $\mathbf{R}$, computed from the target camera poses $\{c^v_t\}$, encode the geometric relationship between each target viewpoint and the scene.

PRISM builds upon LYRA~\cite{bahmani2025lyra}, which reconstructs a 3D scene from $\mathbf{I}$ by distilling a camera-controlled video diffusion model Gen3C~\cite{ren2025gen3c} into a feed-forward 3DGS decoder. Specifically, Gen3C generates videos $\mathbf{X} = \{\mathbf{x}^v\}_{v=1}^{V}$ for all $V = 6$ trajectories ($L = 121$ frames each) through iterative denoising.
% #---# R1: attribute the frozen decoder to LYRA in the main text.
% #---# Folded into the sentence that introduces the decoder.
% A pre-trained Cosmos VAE encoder $\mathcal{E}$ then compresses $\mathbf{X}$ into latents, and the frozen 3DGS decoder $\mathcal{D}$ decodes them into 3D Gaussian primitives:
A pre-trained Cosmos VAE encoder $\mathcal{E}$ then compresses $\mathbf{X}$ into latents, and the frozen 3DGS decoder $\mathcal{D}$, adopted from LYRA, decodes them into 3D Gaussian primitives:
\begin{equation}
    \mathbf{Z} = \mathcal{E}(\mathbf{X}), \quad \mathcal{G} = \mathcal{D}(\mathbf{Z}, \mathbf{R})
    \label{eq:lyra_pipeline}
\end{equation}
where $\mathbf{Z} = \{\mathbf{z}^v\}_{v=1}^{V}$ with each $\mathbf{z}^v \in \mathbb{R}^{T \times C \times h \times w}$, $T=16$, $C=16$, $h=H/8$, $w=W/8$.
The Cosmos VAE temporally compresses $L$ raw frames to $T=16$ latent frames (compression ratio ${\sim}7.5{\times}$).
% Gen3C requires approximately 2.8 hours, as its denoising process across $V \times L = 726$ frames dominates the total inference time.
Gen3C requires approximately 2.8 hours\footnote{Measured using the official LYRA codebase on the same A6000 GPU, at $\sim$1{,}668 seconds per camera trajectory with $L=121$ frames, which totals $\sim$2.8 hours for $V=6$ trajectories.} per scene for new view synthesis, as its denoising process across $V \times L = 726$ frames dominates the total inference time.
As illustrated in Fig.~\ref{fig:framework}, PRISM addresses this bottleneck with a fast feed-forward encoder that approximates $\mathbf{Z}$ without iterative diffusion.
The intermediate features of the decoder's cross-view attention layers, which aggregate information across all $V$ views and encode geometric relationships that generalize across scenes, serve as distillation targets during training.

\begin{figure*}[tbp]
    \centering
    \includegraphics[width=0.8\linewidth]{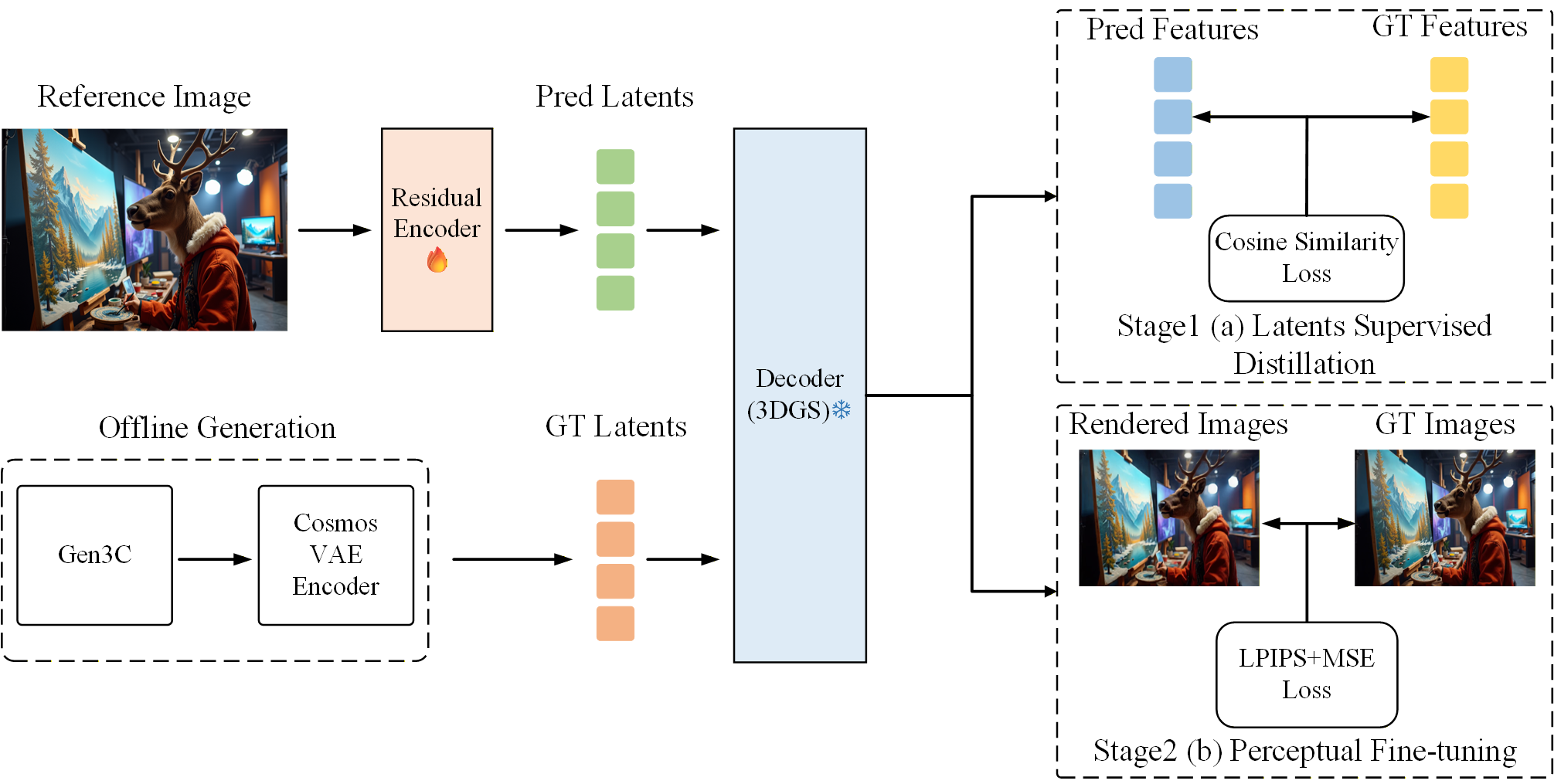}
    \caption{\textbf{Two-stage training pipeline of PRISM.} The residual encoder predicts latents from the reference image (see Fig.~\ref{fig:framework} for encoder input details), while Ground Truth (GT) latents are generated offline via Gen3C and the Cosmos VAE encoder. Both sets of latents are passed through the frozen 3DGS decoder. In Stage 1~(a), intermediate decoder features are aligned via cosine similarity loss. In Stage 2~(b), rendered images are compared against GT images via an LPIPS and MSE loss.}
    \label{fig:training}
    \vspace{-0.3cm}
\end{figure*}

\subsection{Geometric Warp-Residual Decomposition}
\label{sec:decomposition}
 
\noindent\textbf{Geometric decomposition.}
A naive approach is to train the encoder to predict $\hat{\mathbf{Z}}$ directly from $\mathbf{I}$ and $\mathbf{R}$, but this requires the encoder to generate all target-view content without any geometric guidance, which places an unreasonably high burden on a lightweight model.
 
Given monocular depth and target camera poses, forward warping can cover up to $\sim$90\% of a target view directly from the input, as illustrated in the previous Fig.~\ref{fig:illustration}, with uncovered regions comprising disoccluded areas and depth estimation errors.
This observation motivates a decomposition of the prediction into two components: a parameter-free geometric prior $\mathbf{z}^v_{\text{blend},t}$ and a learned residual correction $\Delta_{\theta,t}^v$ that accounts for the remaining low-coverage regions:
\begin{equation}
    \hat{\mathbf{z}}^v_t = \mathbf{z}^v_{\text{blend},t} + \Delta_{\theta,t}^v
    \label{eq:residual}
\end{equation}
Since the invalid or unreliable warped regions occupy a small fraction of the latent grid on average, the encoder only needs to correct these locations. This makes the prediction problem substantially more tractable for a lightweight model.
Furthermore, the residual formulation provides a natural inductive bias: the encoder has little incentive to deviate from the geometric prior in well-covered regions and instead concentrates its corrections on disoccluded areas.
 
\noindent\textbf{Geometry-guided forward warping.}
We encode $\mathbf{I}$ with the Cosmos VAE to obtain the reference latent $\mathbf{z}_{\text{ref}} = \mathcal{E}(\mathbf{I}) \in \mathbb{R}^{C \times h \times w}$, and estimate monocular depth $\mathbf{D} \in \mathbb{R}^{H \times W}$ with MoGe~\cite{wang2025moge}.
We perform z-buffered forward splatting directly in the Cosmos latent space.
For each target camera pose $c^v_t$, every reference pixel $(i, j)$ is unprojected to a 3D point via $\mathbf{D}$, transformed to the target camera frame, and reprojected onto the target image plane, where the latent value $\mathbf{z}_{\text{ref}}(i, j)$ is deposited at the corresponding target location.
When multiple source pixels project to the same target location, z-buffering retains the value from the nearest source point in camera depth, so that foreground surfaces correctly occlude background content.
The operation produces the warped latent $\mathbf{z}^v_{\text{warp},t} \in \mathbb{R}^{C \times h \times w}$ and a binary validity mask $\mathbf{m}^v_t \in \{0,1\}^{h \times w}$ that marks spatial locations with valid warp coverage.

% The Cosmos VAE applies $8\times$ spatial downsampling, which acts as a low-pass filter that absorbs sub-pixel depth estimation errors and reduces the residual variance the encoder needs to correct, and is one reason why a 26M-parameter encoder suffices where a full diffusion model would otherwise be required.
Warping in this space keeps the prior in the same representation as the supervision and the frozen decoder, avoiding an extra encode-decode round trip per view.
This reduction in residual complexity enables a lightweight feed-forward encoder to suffice at inference without requiring a full diffusion model.
 
Regions without valid warp coverage, caused by occlusion or out-of-field-of-view reprojection, are filled with $\mathbf{z}_{\text{ref}}$ to form the blended prior:
\begin{equation}
    \mathbf{z}^v_{\text{blend},t} = \mathbf{m}^v_t \odot \mathbf{z}^v_{\text{warp},t} + (1 - \mathbf{m}^v_t) \odot \mathbf{z}_{\text{ref}}
    \label{eq:blend}
\end{equation}
where $\odot$ denotes element-wise multiplication.
% #---# R2: explain why disoccluded areas take the reference latent.
% #---# Rationale moved to directly follow the fill choice.
The fused representation $\mathbf{z}_{\text{blend}, t}^v$ will keep the encoder's task bounded to residual correction. $\mathbf{z}_{\text{ref}}$ supplies content specific to the input image, whereas zeros or a learned token are identical regardless of input. The alternative would force $\Delta_{\theta,t}^v$ to synthesize appearance from context alone, reducing to unconstrained generation and undermining the inductive bias above. Since the Cosmos VAE temporally compresses $L$ frames to $T=16$ latent frames, $\mathbf{z}_{\text{ref}}$ has a single temporal dimension. We broadcast it across all $T$ target latent frames, so that each temporal frame receives a dedicated geometric prior warped to its corresponding camera pose.

\noindent\textbf{Residual encoder.}
% The encoder infers the content of disoccluded regions from the surrounding warped context, corrects geometric distortions near depth discontinuities, and maintains multi-view consistency across all $V$ predicted latent volumes.
The encoder infers the content of disoccluded regions from the surrounding warped context and corrects geometric distortions near depth discontinuities. In addition, it maintains multi-view consistency across all $V$ predicted latent volumes.
 
To address this, we design a lightweight transformer encoder $f_\theta$ that takes the warped latents, validity masks, reference latent, and Pl\"{u}cker ray embeddings as a multi-channel concatenated input and predicts the residual corrections:
\begin{equation}
    \{\Delta_{\theta,t}^v\} = f_\theta\!\left(\{\mathbf{z}^v_{\text{warp},t}\}, \{\mathbf{m}^v_t\}, \mathbf{z}_{\text{ref}}, \mathbf{R}\right)
    \label{eq:encoder}
\end{equation}
% #---# R2: give a full encoder specification.
% #---# Compact spec here, full table in Supp. Sec. 1; factorized attention.
$f_\theta$ consists of 8 transformer blocks of width 512 with factorized attention, alternating between PerView, CrossView, and Temporal attention rather than joint attention over all $V$ $\times$ $T$ tokens (Supp.~Sec.~1).
The encoder holds 26M parameters, alongside the 33M of the frozen decoder it feeds.
Each input channel serves a distinct role: the warped latent $\mathbf{z}^v_{\text{warp},t}$ provides the geometric prior, and the validity mask $\mathbf{m}^v_t$ explicitly signals to the encoder which regions require correction.
The reference latent $\mathbf{z}_{\text{ref}}$ supplies high-frequency texture context for disoccluded regions, and the Pl\"{u}cker embeddings $\mathbf{R}$ encode the relative viewpoint geometry between the input and each target camera.

The encoder processes all $V$ views jointly rather than independently.
This joint processing is essential for geometric consistency: independently predicted residuals for different trajectories can produce mutually inconsistent latent volumes when decoded by the 3DGS decoder $\mathcal{D}$.
The decoder expects cross-view coherent latents to reconstruct a globally consistent 3D scene.
By attending across views, the encoder can leverage geometric constraints, such as the fact that overlapping regions across trajectories should decode to the same 3D Gaussians, to produce coherent residuals.
At inference, the predicted latents $\hat{\mathbf{Z}}$ are decoded directly by $\mathcal{D}$ without Gen3C.

\subsection{Scene-Agnostic Training Objectives}
\label{sec:training}
 
The two-stage training pipeline is illustrated in Fig.~\ref{fig:training}.
During training, Gen3C generates target videos $\mathbf{X}^*$ for each scene offline, and the corresponding GT latents $\mathbf{Z}^* = \mathcal{E}(\mathbf{X}^*)$ serve as supervision for both stages.
 
\noindent\textbf{Stage 1: Latents Supervised Distillation.}
A naive approach is to directly supervise $\hat{\mathbf{Z}}$ against $\mathbf{Z}^*$ with an $\ell_2$ loss.
% However, direct latent supervision leads to scene-specific memorization: since Gen3C latents carry scene-specific appearance statistics, $\ell_2$ supervision encourages the encoder to fit these statistics precisely rather than developing a generalizable prediction strategy, and the resulting model fails to generalize to unseen scenes.
However, direct latent supervision leads to scene-specific memorization.
Since Gen3C latents carry scene-specific appearance statistics, $\ell_2$ supervision encourages the encoder to fit these statistics precisely rather than developing a generalizable prediction strategy, and the resulting model fails to generalize to unseen scenes.
 
To address this, we supervise the encoder through the intermediate features of the frozen 3DGS decoder $\mathcal{D}$ rather than the raw latent values.
We specifically target the cross-view attention layers of $\mathcal{D}$, as these are the only layers in $\mathcal{D}$ that exchange information across all $V$ views simultaneously.
Their representations therefore encode multi-view geometric consistency rather than per-view appearance, which makes them well-suited as distillation targets for a model that needs to generalize across diverse scenes.
As shown in Fig.~\ref{fig:training}(a), both $\hat{\mathbf{Z}}$ and $\mathbf{Z}^*$ are passed through $\mathcal{D}$, and features are extracted at the selected blocks $\mathcal{B}$:
\begin{equation}
    \mathcal{L}_{\text{feat}} = \frac{1}{|\mathcal{B}|} \sum_{l \in \mathcal{B}} \left(1 - \text{CosSim}\!\left(\phi_l(\hat{\mathbf{Z}}),\, \phi_l(\mathbf{Z}^*)\right)\right)
    \label{eq:feat_loss}
\end{equation}
where $\phi_l(\cdot)$ denotes the feature map at block $l$.
Cosine similarity is chosen over $\ell_2$ distance for two reasons.
First, it is scale-invariant and matches feature directions rather than magnitudes; this prevents the encoder from memorizing scene-specific activation scales.
Second, feature directions encode the geometric structure that the decoder relies on for consistent 3D reconstruction, which makes cosine similarity a more geometrically meaningful objective than absolute feature distance.
 
\noindent\textbf{Stage 2: Perceptual Fine-tuning.}
While latents supervised model distillation establishes geometric generalization, feature-space alignment does not guarantee perceptual rendering quality. For instance, two latents can have similar decoder features yet produce rendered images that differ substantially in fine-grained texture and sharpness.
To bridge this gap, we fine-tune the encoder from the Stage 1 checkpoint with a perceptual loss computed on rendered images through the frozen 3DGS decoder, as shown in Fig.~\ref{fig:training}(b):
\begin{equation}
    \mathcal{L}_{\text{render}} = \lambda_{\text{LPIPS}}\, \mathcal{L}_{\text{LPIPS}}(\hat{\mathbf{I}}, \mathbf{I}^*) + \lambda_{\text{MSE}}^{\text{dyn}}\, \mathcal{L}_{\text{MSE}}(\hat{\mathbf{I}}, \mathbf{I}^*)
    \label{eq:render_loss}
\end{equation}
where $\hat{\mathbf{I}}$ are the rendered novel views and $\mathbf{I}^*$ are the GT images rendered from target latents.
Note that $\mathbf{I}^*$ is rendered from $\mathbf{Z}^*$ through the frozen decoder rather than from real captured frames: PRISM distills the LYRA rendering pipeline, so a more consistent supervision target is LYRA's rendered output rather than pixel-level ground truth.
LPIPS captures perceptual similarity at the feature level, while MSE penalizes low-frequency brightness and color discrepancies.
The dynamic weight $\lambda_{\text{MSE}}^{\text{dyn}}$ is adjusted during training to maintain a fixed contribution ratio of $0.5$ relative to the LPIPS term, so that neither loss dominates.
The two-stage design decouples geometric generalization from perceptual quality alignment: Stage 1 establishes a generalizable geometric initialization, and Stage 2 refines rendered appearance from that foundation.

\begin{table*}[htbp]
\centering
\caption{\textbf{Quantitative comparison} on RealEstate10K, DL3DV, and Tanks-and-Temples. \textbf{Bold} indicates best, \underline{underline} indicates second best. Baseline numbers are as published, under their own protocols.}
\label{tab:main}
\resizebox{\textwidth}{!}{%
\begin{tabular}{ccccccccccc}
\toprule
\multirow{2}{*}{Method} & \multicolumn{3}{c}{RealEstate10K} & \multicolumn{3}{c}{DL3DV} & \multicolumn{3}{c}{Tanks-and-Temples} & \multirow{2}{*}{Time} \\
\cmidrule(lr){2-4} \cmidrule(lr){5-7} \cmidrule(lr){8-10}
& PSNR$\uparrow$ & SSIM$\uparrow$ & LPIPS$\downarrow$ & PSNR$\uparrow$ & SSIM$\uparrow$ & LPIPS$\downarrow$ & PSNR$\uparrow$ & SSIM$\uparrow$ & LPIPS$\downarrow$ & \\
\midrule
ZeroNVS~\cite{sargent2024zeronvs}    & 13.01 & 0.378 & 0.448 & 13.35 & 0.339 & 0.465 & 12.94 & 0.325 & 0.470 & $\sim$3h \\
ViewCrafter~\cite{yu2025viewcrafter}  & 16.84 & 0.514 & 0.341 & 15.53 & 0.525 & 0.352 & 14.93 & 0.483 & 0.384 & $\sim$6min \\
Wonderland~\cite{liang2025wonderland} & 17.15 & 0.550 & 0.292 & 16.64 & 0.574 & \underline{0.325} & 15.90 & 0.510 & \underline{0.344} & $\sim$5min \\
LYRA~\cite{bahmani2025lyra}           & \textbf{21.79} & \textbf{0.752} & \textbf{0.219} & \textbf{20.09} & \underline{0.583} & \textbf{0.313} & \textbf{19.24} & \textbf{0.570} & \textbf{0.336} & $\sim$2.8h \\
\midrule
PRISM (ours)                          & \underline{21.00} & \underline{0.741} & \underline{0.264} & \underline{19.52} & \textbf{0.622} & 0.347 & \underline{18.29} & \underline{0.560} & 0.369 & $\sim$36s \\
\bottomrule
\end{tabular}%
}
\vspace{-0.1cm}
\end{table*}

% Inference time is measured per scene on a single A6000 GPU, covering 6 synthetic trajectories of 121 frames each.

\section{Experiments}
\label{sec:exp}

\subsection{Experimental Setup}
\label{sec:setup}

\noindent\textbf{Datasets and baselines.}
We evaluate PRISM on three benchmarks: RealEstate10K~\cite{zhou2018stereo} (RE10K), DL3DV~\cite{ling2024dl3dv}, and Tanks-and-Temples~\cite{knapitsch2017tanks} (T\&T), reporting PSNR, SSIM, and LPIPS on all three at $704 \times 1280$ resolution.
We sample frames at a stride of 2, 1, and 3 respectively, taking 14 targets per clip, over 300 RE10K scenes, 140 DL3DV scenes, and 28 T\&T clips.
We compare against a variety of baselines including ZeroNVS~\cite{sargent2024zeronvs}, ViewCrafter~\cite{yu2025viewcrafter}, Wonderland~\cite{liang2025wonderland}, and LYRA~\cite{bahmani2025lyra}.
% #---# R2: evaluate LYRA under our own protocol, not literature numbers.
% #---# Merged with the no-source-code claim it contradicted.
% Since no source code is publicly available to evaluate these methods, baseline results are taken directly from the respective papers and may reflect different evaluation protocols.
Since only LYRA releases public code, Table~\ref{tab:main} reports every baseline's numbers as stated in its own paper, and we re-evaluate LYRA's full pipeline under our own protocol, alongside a frame-sampling-stride sensitivity analysis in Supp.~Sec.~6.

\noindent\textbf{Implementation details.}
PRISM is trained on 401 synthetic scenes from the LYRA dataset~\cite{bahmani2025lyra}, generated from diverse text prompts, with no real-world images or multi-view captures.
Notably, all three evaluation benchmarks are unseen during training.
Gen3C~\cite{ren2025gen3c} provides multi-view latent supervision across 6 trajectories of $L=121$ frames per scene.
The residual encoder is first trained with the latents supervised distillation loss at a learning rate of $1 \times 10^{-4}$ for 5k steps, then fine-tuned from the Stage 1 checkpoint with the perceptual rendering loss at $1 \times 10^{-4}$ for another 10k steps, each with a batch size of 1.
All experiments are conducted on a single NVIDIA A6000 GPU.

\subsection{Main Results}
\label{sec:results}

Table~\ref{tab:main} reports quantitative comparisons on three benchmarks.
PRISM learns from LYRA~\cite{bahmani2025lyra}, whose latents supervise training, so LYRA sets the ceiling these numbers are read against. PRISM achieves competitive reconstruction quality on RE10K and DL3DV. Specifically, it reaches an SSIM of 0.622 on DL3DV that ranks first among all methods and takes just 36 seconds per scene, exhibiting a $277\times$ speedup. PRISM stays within 0.8 dB of LYRA on RE10K and DL3DV; the gap widens on T\&T. Re-evaluating LYRA under our protocol reproduces this ordering on T\&T, while on RE10K it places PRISM ahead on PSNR and SSIM, at 24 minutes against 112 hours (Supp.~Sec.~6).
% #---# R2: as above. Drops the absolute "strongest result" claim and
% #---# reports the matched-protocol reversal from Supp. Sec. 6.
% Furthermore, on the more challenging T\&T benchmark, PRISM outperforms LYRA across all three metrics, the strongest result among all evaluated methods.

% #---# Consistency: "superior" sat just below its own negation.
% Against other diffusion-based methods, PRISM outperforms both ViewCrafter~\cite{yu2025viewcrafter} and Wonderland~\cite{liang2025wonderland} across all three benchmarks. This indicates that geometric warp-residual decomposition achieves superior scene reconstruction quality without iterative sampling at inference.
Against other diffusion-based methods, PRISM leads ViewCrafter~\cite{yu2025viewcrafter} on every metric and Wonderland~\cite{liang2025wonderland} on PSNR and SSIM. This indicates that geometric warp-residual decomposition remains competitive with generative approaches without diffusion sampling at inference.

\begin{table}[tbp]
\centering
\caption{\textbf{Ablation study} of each component on RE10K dataset.}
\label{tab:ablation}
\begin{tabular}{lccc}
\toprule
Method & PSNR$\uparrow$ & SSIM$\uparrow$ & LPIPS$\downarrow$ \\
\midrule
w/o Residual Encoder     & 19.51 & 0.731 & 0.457 \\
w/o Stage 1 Pre-training & 20.85 & 0.747 & 0.379 \\
w/o Stage 2 Fine-tuning  & 21.22 & \textbf{0.766} & 0.269 \\
\midrule
PRISM (ours)             & \textbf{21.28} & 0.763 & \textbf{0.242} \\
\bottomrule
\end{tabular}
\end{table}

\begin{table}[tbp]
\centering
\caption{\textbf{Latents distillation layer analysis} on RE10K dataset, measured after Stage 1.}
\label{tab:layer_analysis}
\begin{tabular}{lccc}
\toprule
Blocks $\mathcal{B}$ & PSNR$\uparrow$ & SSIM$\uparrow$ & LPIPS$\downarrow$ \\
\midrule
$\{15\}$                   & 21.14          & 0.762          & 0.272          \\
$\{7, 15\}$                & \textbf{21.22} & 0.766          & \textbf{0.269} \\
$\{0, 7, 15\}$             & \textbf{21.22}          & \textbf{0.768} & 0.273          \\
$\{0, 7, 14, 15\}$         & 21.13          & 0.766          & 0.270          \\
\bottomrule
\end{tabular}
% \vspace{-0.3cm}
\end{table}

\subsection{Ablation Studies}
\label{sec:ablation}

To assess the contribution of each design choice, we conduct ablation studies on a randomly subsampled 40-scene RE10K split at stride 2, as shown in Table~\ref{tab:ablation}.
 
First, when the residual encoder is removed, PSNR drops by 1.77 dB and LPIPS degrades from 0.242 to 0.457, which reveals that the geometric warp prior alone cannot recover fine-grained appearance detail. Moreover, without Stage 1 pre-training, both PSNR and LPIPS degrade substantially, suggesting that feature distillation provides a stable initialization that Stage 2 depends on.
Furthermore, Stage 1 alone achieves the highest SSIM and strong PSNR, but Stage 2 is necessary to close the perceptual quality gap (LPIPS 0.269 vs.\ 0.242), at a marginal SSIM cost.
The full design achieves the best PSNR of 21.28 dB and LPIPS of 0.242, with each stage contributing to reconstruction quality.

\subsection{Qualitative Visualizations}
\label{sec:qual}

\begin{figure}[t]
    \centering
    \includegraphics[width=1.0\linewidth]{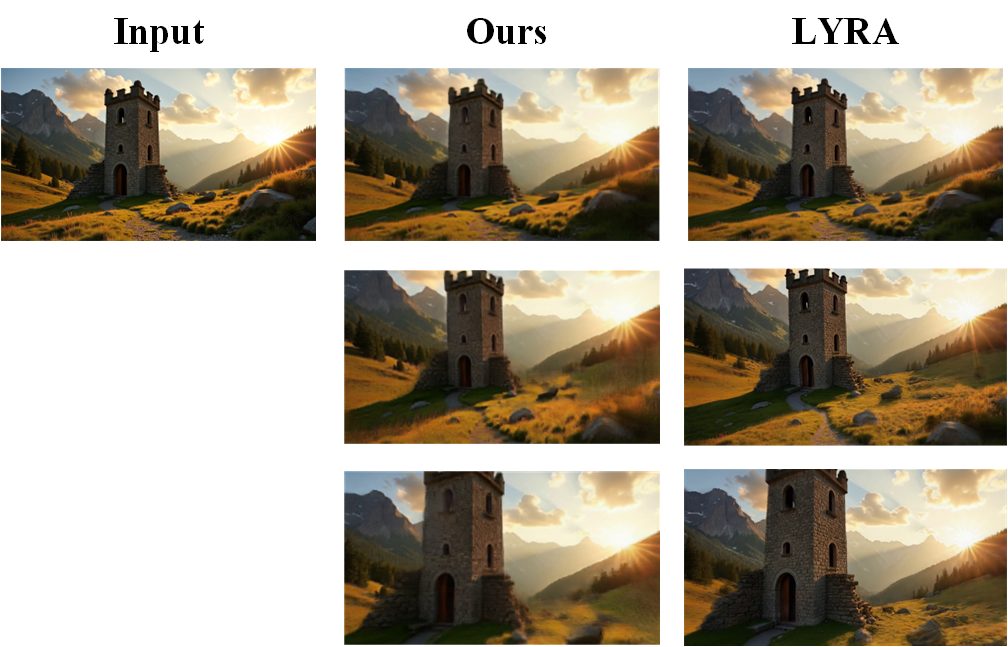}
    \caption{\textbf{Qualitative comparison on a synthetic scene.} Given the input image (left), PRISM (middle) and LYRA~\cite{bahmani2025lyra} (right) synthesize three novel viewpoints. PRISM produces visually comparable results to LYRA without diffusion sampling at inference.}
    \label{fig:lyra-visual}
    % \vspace{-0.3cm}
\end{figure}

On a synthetic scene from the training distribution (Fig.~\ref{fig:lyra-visual}), PRISM matches LYRA across all three viewpoints in geometry, lighting, and texture.

\begin{figure*}[htbp]
    \centering
    \includegraphics[width=0.84\linewidth]{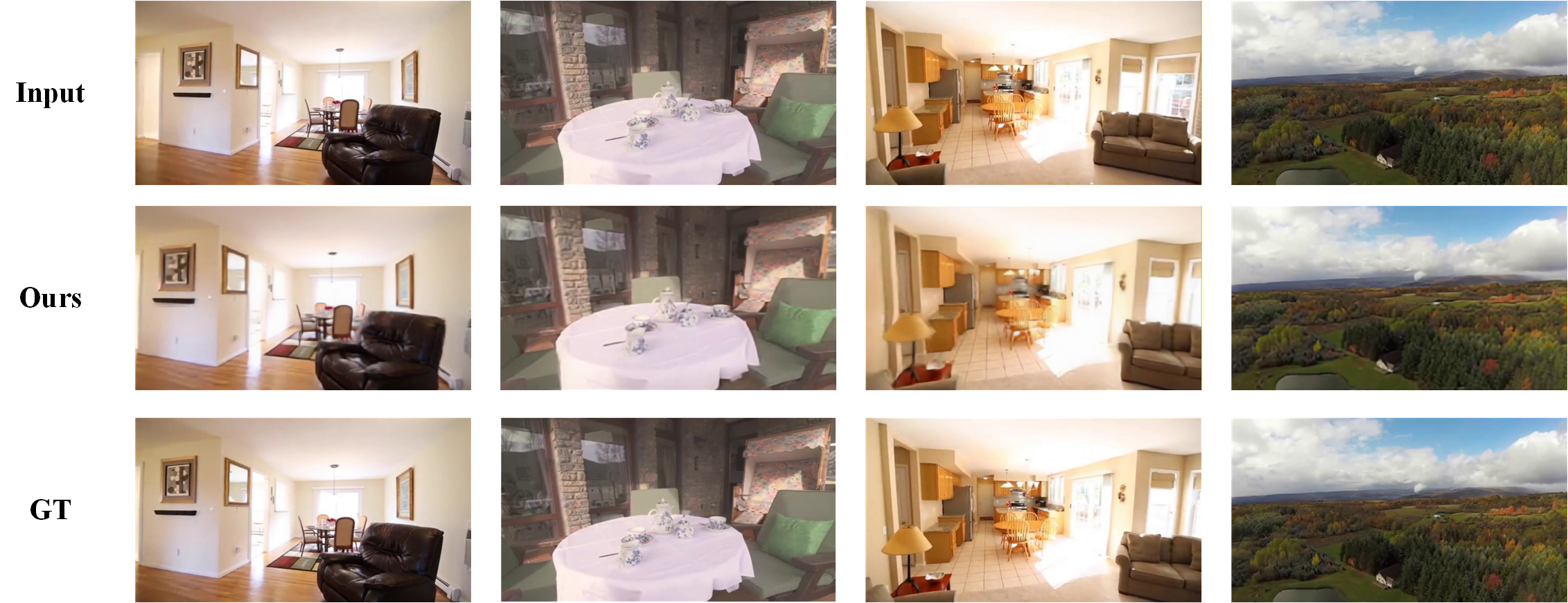}
    \caption{\textbf{Qualitative results on real-world scenes from RE10K.} Each column shows a different scene. Frame 14 of each 14-frame clip is shown for comparison. Rows from top to bottom: input image (Frame 1), PRISM render (Frame 14), and ground truth (Frame 14). PRISM generalizes to real-world scenes despite being trained exclusively on synthetic data.}
    \label{fig:real-visual}
    \vspace{-0.1cm}
\end{figure*}

As shown in Fig.~\ref{fig:real-visual}, we further evaluate on five real-world scenes from RE10K.
Across diverse indoor and outdoor settings, PRISM outputs consistently align with ground truth in structure, color, and scene layout.
The warp-residual decomposition therefore generalizes beyond the synthetic training domain, producing coherent reconstructions from a single image of an unseen real-world scene. Supp.~Sec.~6 places PRISM and LYRA side by side on real RE10K scenes at low-coverage viewpoints, where both soften in the same disoccluded regions.

\subsection{Latents Distillation Layer Analysis}

The frozen 3DGS decoder consists of 16 layers, where blocks 7 and 15 are the only two cross-view Transformer layers; the remaining 14 are Mamba-2 blocks.
To analyze which layers serve as the most effective distillation targets, we ablate the choice of $\mathcal{B}$ after Stage 1 on the same 40-scene RE10K split used in Sec.~4.3.

Table~\ref{tab:layer_analysis} shows that the configuration $\{7, 15\}$ (Row 2) achieves the best PSNR and LPIPS among all choices, suggesting that both cross-view attention layers contribute to geometric generalization.
When Mamba-2 layers are added to the distillation target, no improvement follows, as Row 3 degrades LPIPS relative to Row 2 despite a marginal SSIM gain, and Row 4 further reduces PSNR.
These results indicate that cross-view attention layers are the primary source of generalizable geometric supervision, and that Mamba-2 features introduce noise as distillation targets.

\subsection{Efficiency Analysis}
\label{sec:efficiency}

\begin{figure}[tbp]
    \centering
    \includegraphics[width=1.0\linewidth]{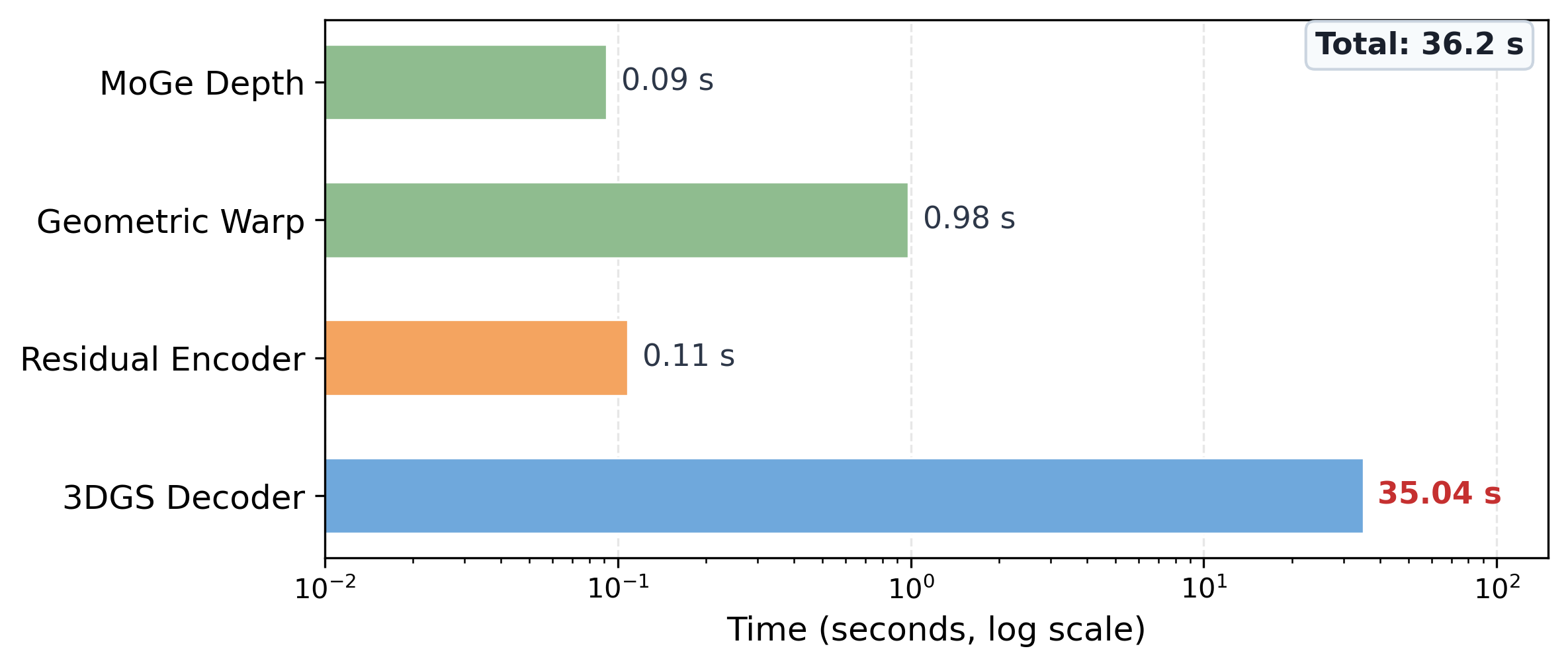}
    \caption{\textbf{Per-component inference time} breakdown of PRISM.}
    \label{fig:latency}
    \vspace{-0.3cm}
\end{figure}

As illustrated in Fig.~\ref{fig:latency}, the average inference time of PRISM is 36.2 seconds per scene.
The per-component breakdown reveals that the 3DGS decoder dominates at 35.04 seconds, accounting for 97\% of total inference time.
In contrast, PRISM's three new components (MoGe depth estimation 0.09s, geometric warp 0.98s, and residual encoder 0.11s) together contribute only 1.18 seconds.
% This suggests that future work of accelerating the 3DGS Decoder would directly compound PRISM's existing $277\times$ speedup.
The bottleneck has shifted away from diffusion sampling. 
% Future acceleration of the 3DGS decoder would directly compound the existing $277\times$ speedup.

\subsection{Coverage and Scene-Complexity Analysis}
\label{sec:analysis}

\begin{figure}[tbp]
    \centering
    \includegraphics[width=1.0\linewidth]{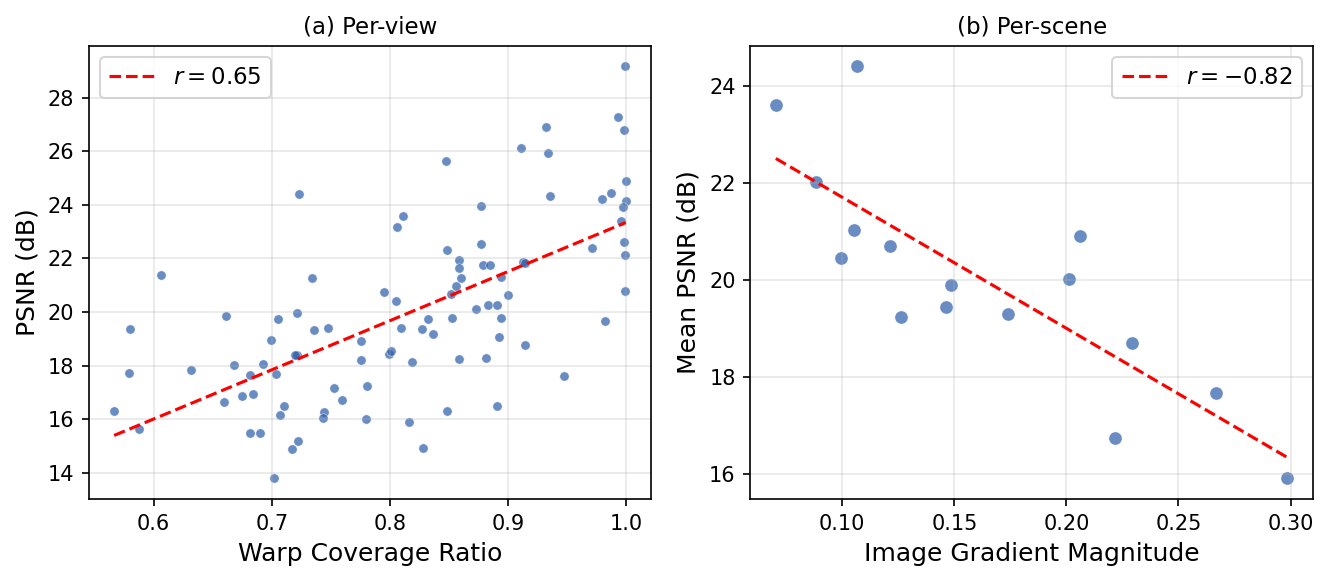}
    \caption{\textbf{Coverage and scene-complexity correlation} on the LYRA dataset. (a) Per-view warp coverage versus PSNR. (b) Per-scene image gradient magnitude versus mean PSNR.}
    \label{fig:coverage}
    \vspace{-0.3cm}
\end{figure}

We analyze how warp coverage and scene complexity relate to reconstruction quality on the sub-sampled 16 test scenes from the LYRA dataset~\cite{bahmani2025lyra}.
As shown in Fig.~\ref{fig:coverage}(a), per-view warp coverage correlates positively with PSNR ($r=0.65$).
Views with higher coverage leave less content for the residual encoder to correct, which aligns with the decomposition introduced in Sec.~\ref{sec:decomposition}.
Moreover, Fig.~\ref{fig:coverage}(b) reveals that per-scene image gradient magnitude correlates negatively with mean PSNR ($r=-0.82$).
This suggests that scenes with richer high-frequency content present greater reconstruction challenges for the residual encoder.
Together, these observations indicate that quality variation in PRISM is geometry- and content-driven rather than random. The residual encoder concentrates on low-coverage and high-complexity regions, which is the role it is designed to fill.

\section{Conclusion}
\label{sec:conclu}

In this work, we present PRISM, a feed-forward framework for single-image 3D reconstruction that shows diffusion sampling supervises training but is not needed at inference to synthesize competitive novel views.
PRISM separates what geometry can already explain from what remains uncertain, which allows a lightweight feed-forward encoder to replace iterative diffusion sampling at inference.
The two-stage training strategy transfers a model trained without real-world supervision to real scenes: latents supervised distillation aligns the encoder with the frozen 3DGS decoder, and perceptual fine-tuning closes the appearance gap.
PRISM matches geometry-based reconstruction in efficiency while staying competitive in fidelity.

% \section*{LLM Usage Statement}
% This work made use of Claude (Anthropic), accessed via Claude Code. The tool generated and revised parts of the text and of the code, all of which the authors reviewed, edited and approved. It also assisted in preparing draft layouts that served as references for figures drawn by the authors, in analyzing experimental results and in reviewing related literature. No LLM was involved in creating or annotating data. All research questions, method designs, experiments and reported results originate from the authors, who bear full responsibility for the paper.
{
    \small
    \bibliographystyle{ieeenat_fullname}
    \bibliography{main}
}


\clearpage
\twocolumn[
  \begin{center}
    {\LARGE \textbf{Supplementary Material}}
    \vspace{1em}
  \end{center}
]

\setcounter{section}{0}
\setcounter{figure}{0}
\setcounter{table}{0}

\section{Architecture Details}

\noindent\textbf{Input representation.}
The residual encoder $f_\theta$ takes a 65-channel concatenated tensor as input per spatial location.
Table~\ref{tab:supp_input} summarizes the channel composition.
The forward-warped reference latent $\mathbf{z}^v_{\text{warp},t}$ (channels 0--15) provides the geometric prior at each target view and frame.
The warp validity mask $\mathbf{m}^v_t$ (channel 16) explicitly indicates which spatial locations have reliable coverage and which require residual correction.
The reference latent $\mathbf{z}_{\text{ref}}$ (channels 17--32) is broadcast across all views and frames to supply appearance and texture details for disoccluded regions.
The Pl\"{u}cker ray embeddings $\mathbf{R}$ (channels 33--64) encode the relative viewpoint geometry between the input camera and each target camera pose.

\begin{table}[h]
\centering
\caption{65-channel input composition of the residual encoder.}
\label{tab:supp_input}
\begin{tabular}{cll}
\toprule
Channels & Content & Shape \\
\midrule
0--15   & Warped latent $\mathbf{z}^v_{\text{warp},t}$ & $16 \times h \times w$ \\
16      & Warp mask $\mathbf{m}^v_t$                    & $1 \times h \times w$  \\
17--32  & Reference latent $\mathbf{z}_{\text{ref}}$    & $16 \times h \times w$ \\
33--64  & Pl\"{u}cker embeddings $\mathbf{R}$           & $32 \times h \times w$ \\
\bottomrule
\end{tabular}
\end{table}

\noindent\textbf{Encoder architecture.}
The residual encoder adopts a ViT-style design with a lightweight attention block schedule.
The 65-channel input is tokenized via a Conv2d patch embedding with kernel size $4 \times 4$ and stride 4, which reduces the spatial resolution from $88 \times 160$ to $22 \times 40 = 880$ tokens per view-frame.
Each token is projected to a hidden dimension of 512, with learned positional embeddings of shape $1 \times 880 \times 512$.

The encoder consists of 8 transformer blocks with 8 attention heads and an MLP ratio of 4.0, arranged in a repeating schedule of PerView $\to$ CrossView $\to$ PerView $\to$ Temporal, repeated twice.
This schedule alternates between spatial reasoning within each view-frame and geometric reasoning across views and time, as detailed in Table~\ref{tab:supp_blocks}.

\noindent\textit{PerView blocks} apply self-attention over the 880 spatial tokens within each view-frame to refine local spatial features without cross-view interference.

\noindent\textit{CrossView blocks} apply self-attention over the $V=6$ view tokens at each spatial position and temporal frame, enabling the encoder to maintain multi-view geometric consistency.

\noindent\textit{Temporal blocks} apply self-attention over the $T=16$ temporal tokens at each spatial position and view, so that predictions remain coherent across frames.

The output of the final block is passed through a linear projection from 512 to 256 dimensions and unpatchified to recover the full spatial resolution, which produces the residual correction $\Delta \in \mathbb{R}^{B \times V \times T \times 16 \times h \times w}$ added to the blended prior $\mathbf{z}^v_{\text{blend},t}$ per Eq.~(3) in the main paper.
The total parameter count of the encoder is approximately 26M.

\begin{table}[h]
\centering
\caption{Block types in the residual encoder and their functions.}
\label{tab:supp_blocks}
\resizebox{\linewidth}{!}{%
\begin{tabular}{lll}
\toprule
Block & Attention scope & Input shape \\
\midrule
PerView   & Spatial self-attn over 880 tokens   & $B \times V \times T \times 880 \times 512$ \\
CrossView & Self-attn over $V=6$ views          & $B \times T \times N \times 6 \times 512$   \\
Temporal  & Self-attn over $T=16$ frames        & $B \times V \times N \times 16 \times 512$  \\
\bottomrule
\end{tabular}}
\end{table}

% #---# R1: frozen-decoder attribution. Here already; also in main Sec 3.1.
\noindent\textbf{3DGS decoder.}
PRISM reuses the frozen 3DGS decoder $\mathcal{D}$ from LYRA~\cite{bahmani2025lyra} without modification; the residual encoder $f_\theta$ is the only trainable component in PRISM.
For completeness, we summarize its architecture here and refer readers to LYRA~\cite{bahmani2025lyra} for further details.
The decoder takes the predicted latents $\hat{\mathbf{Z}}$ and encoded Pl\"{u}cker embeddings as input, each patchified with a $2 \times 2$ patch layer and summed before entering the reconstruction blocks.
It follows the block design of Long-LRM~\cite{ziwen2025longlrm}, where each block consists of one Transformer layer followed by seven Mamba-2 layers, repeated twice for a total of 16 layers with a hidden dimension of 512.
A transposed 3D convolution then maps the hidden representation to 14 Gaussian channels per pixel, which consist of 3D position, scale, rotation quaternion, opacity, and RGB.

% #---# R3: compare with feed-forward methods, incl. parameter counts.
\noindent\textbf{Comparison with feed-forward reconstruction methods.}
Table~\ref{tab:supp_ff_params} compares PRISM's parameter count against representative feed-forward 3D reconstruction methods.
Since DUSt3R, GS-LRM, and VGGT are each a single end-to-end system, the comparable PRISM number is its full pipeline, not the residual encoder alone: a 26M-parameter trainable encoder plus a 33M-parameter decoder reused unmodified from LYRA~\cite{bahmani2025lyra}, for 59M total.
GS-LRM~\cite{zhang2024gslrm} requires 2--4 posed images as input; DUSt3R~\cite{wang2024dust3r} and VGGT~\cite{wang2025vggt} are themselves pose-free but are designed around multiple views, taking image pairs or scaling to hundreds of frames respectively. PRISM instead operates from a single unposed image, and its full pipeline is still an order of magnitude smaller than any of these three, which reflects its narrower task: correcting a geometric prior that warping has already mostly completed, rather than inferring 3D structure from scratch.

\begin{table}[tbp]
\centering
\small
\caption{Parameter count comparison with feed-forward 3D reconstruction methods. PRISM's full pipeline (trainable encoder plus the frozen decoder reused from LYRA) is the comparable full-system number; it is the only method operating from a single image and remains an order of magnitude smaller than any baseline.}
\label{tab:supp_ff_params}
\begin{tabular}{lcc}
\toprule
Method & Params & Input \\
\midrule
DUSt3R~\cite{wang2024dust3r}  & 571M & $\geq$2 unposed images \\
GS-LRM~\cite{zhang2024gslrm}  & 500M & 2--4 posed images \\
VGGT~\cite{wang2025vggt}      & 1.2B & 1+ unposed images \\
PRISM (ours)                  & 59M  & 1 unposed image \\
\bottomrule
\end{tabular}
\end{table}

% #---# Params from each paper; PRISM split from count_params.py on the
% #---# real ckpts. DUSt3R and VGGT are pose-free, GS-LRM needs posed input.

% #---# R2: latent-vs-pixel warping unablated; no non-diffusion baseline.
% #---# Flash3D not cited: its RE10K protocol differs and it omits DL3DV/T&T.
\noindent\textbf{Latent-space warping and non-diffusion baselines.}
Two design choices are not empirically ablated in the current submission.
Warping in the latent space keeps the geometric prior in the same representation as the latent supervision and the frozen decoder, which is why PRISM adopts it.
Flash3D~\cite{szymanowicz20253dv}, discussed in Sec.~2 of the main paper, is the closest single-image, non-diffusion competitor and has public code, but its own reported RE10K protocol (fixed 5/10/$\pm$30-frame view-distance splits) differs from ours and does not cover DL3DV or T\&T; given the same protocol-mismatch risk identified for T\&T (Sec.~6), we avoid citing its literature-reported numbers directly and leave a matched-protocol re-evaluation for future work.

\begin{figure*}[tbp]
    \centering
    \includegraphics[width=1.0\linewidth]{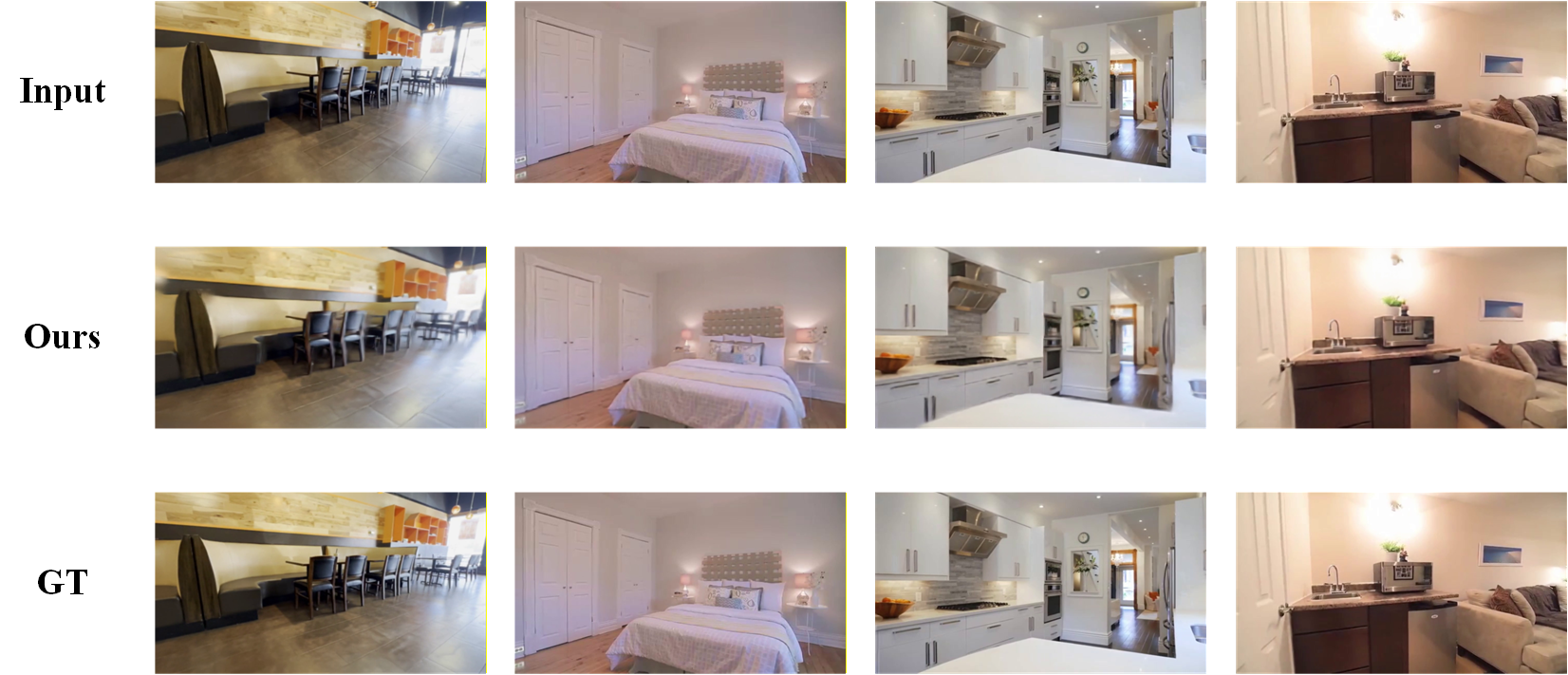}
    \caption{\textbf{Additional qualitative results on DL3DV~\cite{ling2024dl3dv} (column 1) and RE10K~\cite{zhou2018stereo} (columns 2--4).} Each column shows a different scene. Rows from top to bottom: input image (Frame 1), PRISM render (Frame 14), and ground truth (Frame 14).}
    \label{fig:supp_real}
\end{figure*}

\section{Implementation Details}

\noindent\textbf{Training.}
We train PRISM with a warm-up initialization followed by two training stages. All training is conducted on a single NVIDIA A6000 GPU with a batch size of 1.

\noindent\textbf{Initialization.}
The encoder is first pretrained with a pixel-wise latent regression loss on 401 synthetic scenes for 20k steps, which provides a critical initialization for the subsequent feature distillation and perceptual fine-tuning stages.

\noindent\textbf{Stage 1: Latents Supervised Distillation.}
The encoder is initialized from the warm-up checkpoint and trained with the cosine loss at a learning rate of $1 \times 10^{-4}$ for 5k steps on 401 synthetic scenes from LYRA dataset.

\noindent\textbf{Stage 2: Perceptual Fine-tuning.}
The encoder is fine-tuned from the Stage 1 checkpoint with the perceptual rendering loss at $1 \times 10^{-4}$ for another 10k steps.

\noindent\textbf{Inference.}
Our inference time is measured per scene on a single A6000 GPU, covering 6 synthetic trajectories of 121 frames each. 
Specifically, given a single input image $\mathbf{I}$ and target camera poses $\{C_{v,t}\}$, PRISM inference proceeds in four steps.

\noindent\textbf{(1) Reference encoding.}
The Cosmos VAE encoder $\mathcal{E}$ encodes the input image to obtain the reference latent $\mathbf{z}_{\text{ref}} = \mathcal{E}(\mathbf{I}) \in \mathbb{R}^{C \times h \times w}$.

\noindent\textbf{(2) Depth estimation.}
MoGe estimates monocular depth $\mathbf{D}$ at the original image resolution; camera intrinsics are scaled accordingly to the latent resolution with scale factor $s=8$. Because the Cosmos VAE compresses $L=121$ input frames to $T=16$ latent frames, latent frame $t$ of view $v$ takes the camera pose of input frame $v \cdot 121 + 8t$, so each warped latent is paired with the pose of the frame it was compressed from.

\noindent\textbf{(3) Geometric warping.}
For each view $v$ and frame $t$, z-buffered forward splatting warps the reference latent to the target camera pose, producing the warped latent $\mathbf{z}^v_{\text{warp},t}$ and validity mask $\mathbf{m}^v_t$. The blended prior is then formed per Eq.~(3) in the main paper.

\noindent\textbf{(4) Residual prediction and decoding.}
The residual encoder $f_\theta$ predicts the residual correction per Eq.~(4) in the main paper, where $\mathbf{R}$ denotes the Pl\"{u}cker ray embeddings derived from the target camera poses $\{C_{v,t}\}$. The final predicted latent $\hat{\mathbf{z}}^v_t$ is obtained per Eq.~(2), which is decoded by the frozen 3DGS decoder $\mathcal{D}$ to render novel views at arbitrary camera poses.

\section{More Qualitative Results}

We present additional qualitative results on real-world scenes in Fig.~\ref{fig:supp_real}.
Across diverse indoor environments including dining rooms, bedrooms, and kitchens, PRISM produces novel views that closely align with the ground truth in structure, color, and scene layout.

% #---# R4: supplementary videos. Held back for this submission; text below kept for reuse.
% Video comparisons for additional RE10K scenes are included in the supplementary material: PRISM's rendered predictions alongside ground truth. These extend to 60 frames at stride 1, beyond the 14 frames used for Table~1 of the main paper's quantitative results.

% #---# R2: stratify results by per-view warp coverage.
% #---# Same real target frames; only DL3DV spreads enough (0.61-1.00).
% This reuses computed outputs and would show directly how far quality falls
\section{Coverage-Stratified Evaluation}

To examine whether the coverage-quality relationship reported in Sec.~4.7 of the main paper (Fig.~8, measured on the LYRA dataset) holds on an independent benchmark, we stratify PRISM's existing DL3DV evaluation results (140 scenes, Table 1 of the main paper) by per-scene warp coverage.
Coverage is computed against the same real target frames used for PSNR/SSIM/LPIPS, preserving the per-view correspondence used in Fig.~8 rather than the 6 synthetic trajectories used as encoder input.

\begin{table}[h]
\centering
\caption{Coverage-stratified evaluation on DL3DV (140 scenes).}
\label{tab:supp_coverage_dl3dv}
\begin{tabular}{lcccc}
\toprule
Coverage & N & PSNR$\uparrow$ & SSIM$\uparrow$ & LPIPS$\downarrow$ \\
\midrule
$<0.75$      & 5  & 13.16 & 0.344 & 0.587 \\
$0.75$--$0.85$ & 15 & 17.83 & 0.562 & 0.429 \\
$0.85$--$0.90$ & 23 & 18.54 & 0.602 & 0.370 \\
$>0.90$      & 97 & 20.26 & 0.645 & 0.321 \\
\bottomrule
\end{tabular}
\end{table}

Table~\ref{tab:supp_coverage_dl3dv} shows a monotonic degradation across all three metrics as coverage decreases, from 20.26~dB / 0.645 / 0.321 in the highest-coverage bin to 13.16~dB / 0.344 / 0.587 in the lowest.
The degradation is gradual rather than a sharp collapse, consistent with the residual formulation in Sec.~3.2: the encoder's burden grows smoothly as the geometric prior weakens, rather than failing abruptly past some threshold.
This corroborates the coverage-PSNR correlation already reported on the LYRA dataset (Fig.~8, main paper) on an independent, out-of-distribution real-world benchmark.

% #---# R3: analyse failures on reflective/transparent surfaces.
% #---# Four RE10K scenes, reported as found: two near average, two below.

\begin{figure*}[htbp]
    \centering
    \includegraphics[width=0.9\linewidth]{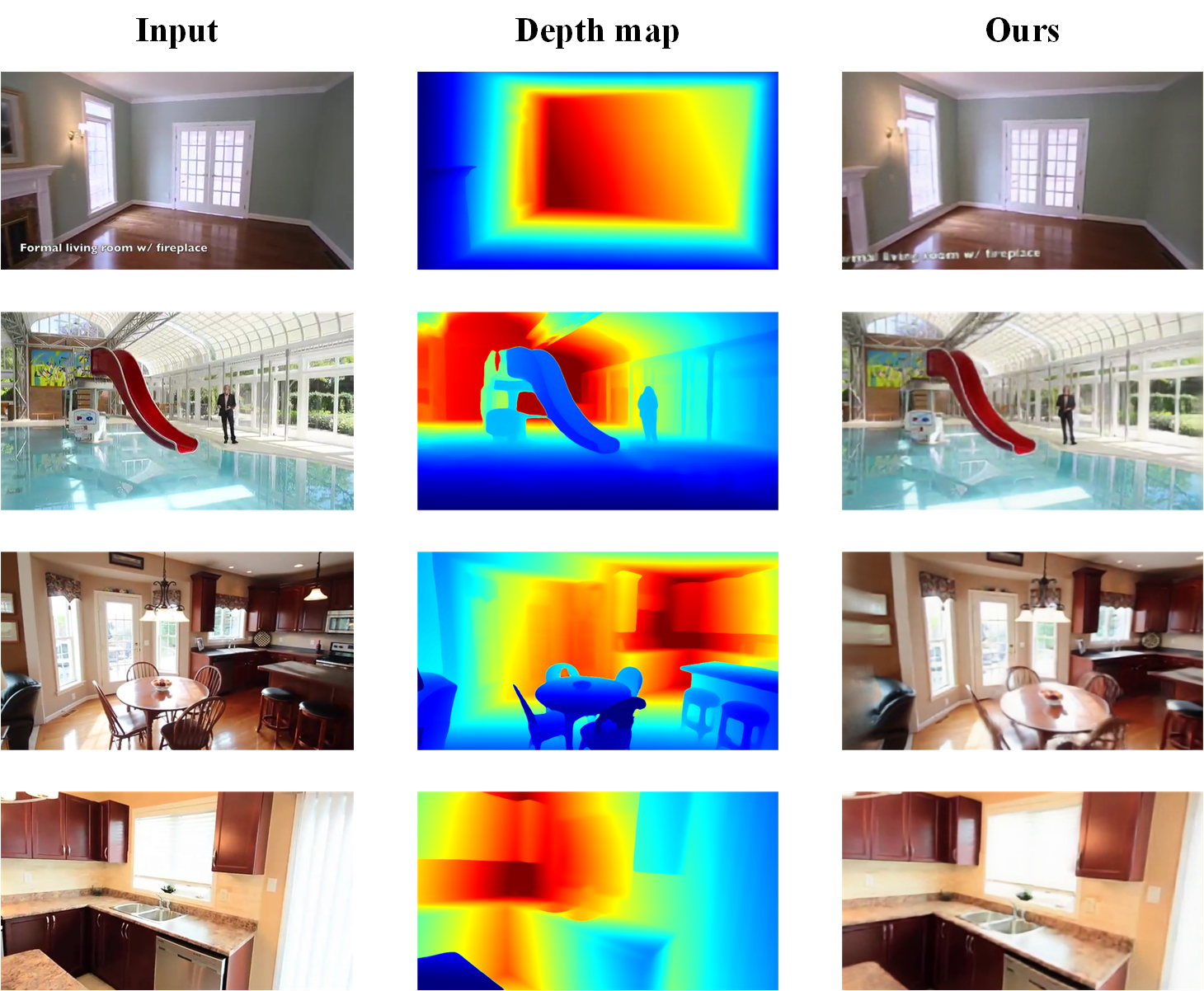}
    \caption{\textbf{PRISM performance on RE10K scenes with reflective/transparent surfaces.} Each row shows a different scene, manually identified as containing prominent reflective or transparent content (glossy floors, glass doors, a glazed pool enclosure, a mirrored sliding door). Columns from left to right: input frame, MoGe depth map, and PRISM render at the furthest evaluated viewpoint. Quantitative results per scene are reported in Table~\ref{tab:supp_reflective}; reconstruction quality varies with the severity of depth ambiguity introduced by these surfaces.}
    \label{fig:supp_reflective}
\end{figure*}

\section{Robustness to Challenging Depth-Estimation Scenes}

To directly evaluate PRISM's robustness to inaccurate depth estimation, we manually identify four RE10K scenes containing prominent reflective or transparent surfaces (glossy hardwood floors, glass doors, a glazed pool enclosure with reflective water, a mirrored sliding door) and report PRISM's performance on each in Table~\ref{tab:supp_reflective}.
A reflective or transparent surface presents an ambiguous signal to a monocular depth estimator: the appearance at that pixel comes from a reflected or transmitted scene rather than from the physical surface itself, so MoGe has no direct visual cue for the surface's true distance.
Since PRISM's geometric prior depends directly on this depth estimate (Sec.~3.2 of the main paper), a depth error at such a surface propagates into the warp, and is corrected only to the extent the residual encoder can compensate for it.

\begin{table}[tbp]
\centering
\small
\caption{Per-scene results on four RE10K scenes with reflective/transparent surfaces, with the full RE10K average for reference. All rows use the stride of 2 adopted in Table~1 of the main paper.}
\label{tab:supp_reflective}
\begin{tabular}{lccc}
\toprule
Scene & PSNR$\uparrow$ & SSIM$\uparrow$ & LPIPS$\downarrow$ \\
\midrule
0b440bf8f7f8a4d0 & 22.67 & 0.695 & 0.254 \\
0a6680fe6e8e09d7 & 22.03 & 0.817 & 0.190 \\
0c24590c68af865f & 20.34 & 0.770 & 0.238 \\
0c209edeb7637dff & 14.76 & 0.559 & 0.445 \\
\midrule
Full RE10K average (N=300) & 21.00 & 0.741 & 0.264 \\
\bottomrule
\end{tabular}
\end{table}

Performance on these four scenes ranges from near or above the RE10K average to well below it, rather than uniformly failing.
The two higher-scoring scenes show that reflective or transparent content alone does not guarantee a large quality drop, including the pool scene, where glazing and reflective water cover a large part of the frame.
The two lower-scoring scenes show the opposite outcome under the same kind of content, indicating that the effect on quality depends on how much the resulting depth error is corrected downstream, not simply on whether such a surface is present.
Fig.~\ref{fig:supp_reflective} shows the corresponding input frames, MoGe depth estimates, and PRISM renders.
For the two lower-scoring scenes, the render is visibly blurred around the reflective regions; for the mirrored sliding-door scene, the depth estimate places that region farther from the camera than the surrounding wall, which is consistent with the ambiguity described above.

% #---# R2: evaluate LYRA under our protocol; covers the T&T stride question.
\section{Matched-Protocol Evaluation}
\label{sec:supp_lyra_matched}

None of ZeroNVS, ViewCrafter, Wonderland, or LYRA release an evaluation script for RE10K, DL3DV, or T\&T (Sec.~4.1 of the main paper), so Table~1 of the main paper reports each baseline's numbers directly from its own paper.
LYRA is the one exception for which this literature-only comparison can be checked independently: although it releases no dedicated evaluation script, its inference code and pretrained checkpoints are public, which lets us re-evaluate its full generative pipeline (Gen3C generation, Cosmos VAE encoding, and the frozen 3DGS decoder) under the exact protocol used for PRISM itself, rather than relying on the number reported in the LYRA paper alone.

\noindent\textbf{Methodology.} For each scene, we run LYRA's official Gen3C pipeline to generate the same six camera trajectories PRISM takes as input. We then denormalize the resulting latents with the mean and standard deviation statistics released alongside the Cosmos tokenizer, and decode them with LYRA's frozen 3DGS decoder.
Rendering, the real target frames, and the optical-flow scale alignment to each dataset's ground-truth poses are identical to PRISM's own evaluation code. The only difference between the two rows of Table~\ref{tab:supp_lyra_matched} is therefore the source of the predicted latents.
A single Gen3C generation takes $\sim$2.8 hours per scene (Sec.~3.1 of the main paper), so for T\&T we evaluate a subsample of $N=8$ scenes (one clip per scene) rather than reproducing PRISM's full protocol. For RE10K, we instead cover the complete $N=40$ scene ablation subset used in Table 2 of the main paper.

\begin{table*}[htbp]
\centering
\caption{Matched-protocol comparison against LYRA's official pipeline, using PRISM's own evaluation code and real target frames. Time is the total wall-clock time to process all scenes/clips in that benchmark; it is reported per benchmark since T\&T and RE10K use different sample sizes $N$. T\&T is evaluated on consecutive frames here, RE10K at stride 2.}
\label{tab:supp_lyra_matched}
\resizebox{0.9\textwidth}{!}{%
\begin{tabular}{lccccccccc}
\toprule
\multirow{2}{*}{Method} & \multicolumn{4}{c}{Tanks-and-Temples ($N=8$)} & \multicolumn{4}{c}{RealEstate10K ($N=40$)} \\
\cmidrule(lr){2-5} \cmidrule(lr){6-9}
 & PSNR$\uparrow$ & SSIM$\uparrow$ & LPIPS$\downarrow$ & Time & PSNR$\uparrow$ & SSIM$\uparrow$ & LPIPS$\downarrow$ & Time \\
\midrule
PRISM & 21.68 & 0.645 & 0.288 & $\sim$4.8 min & 21.28 & 0.763 & 0.242 & $\sim$24 min \\
LYRA (matched protocol) & 22.98 & 0.725 & 0.174 & $\sim$22.4 h & 20.97 & 0.756 & 0.211 & $\sim$112 h \\
\bottomrule
\end{tabular}%
}
\end{table*}

\begin{figure*}[tbp]
    \centering
    \includegraphics[width=1.0\linewidth]{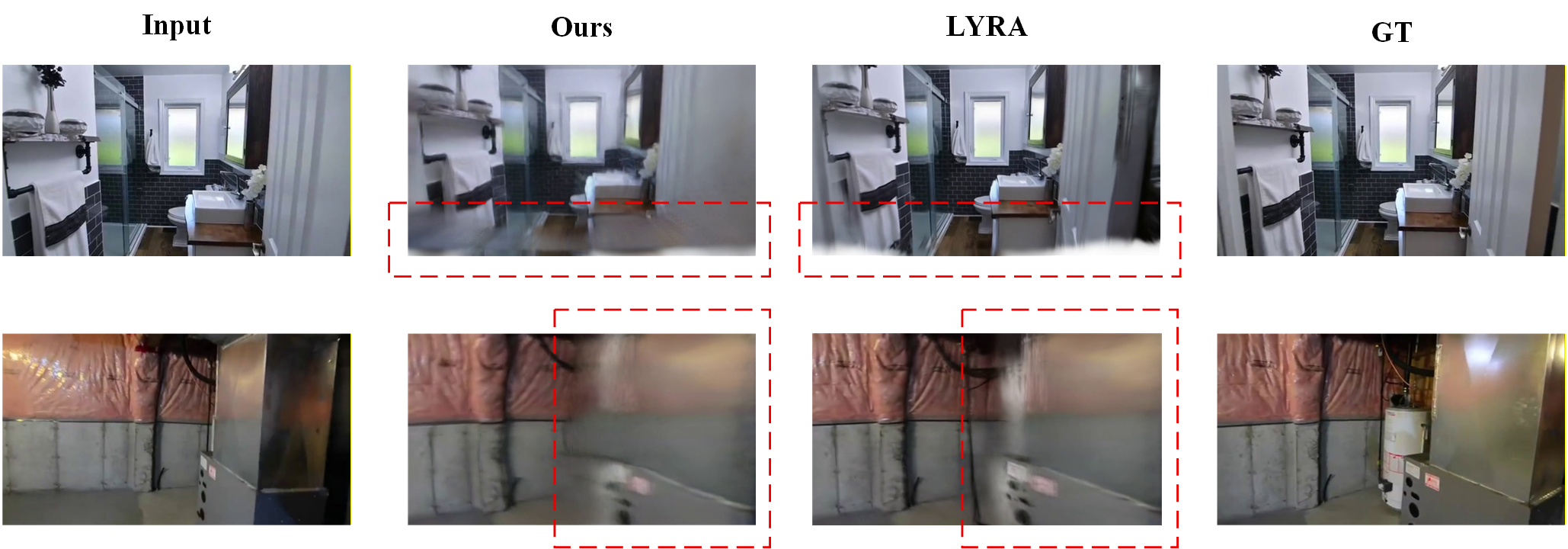}
    \caption{\textbf{Qualitative comparison at low-coverage viewpoints on real RE10K scenes.} Columns: input reference frame, PRISM render, LYRA render (matched protocol), ground truth. Red boxes mark disoccluded regions with reduced sharpness in both PRISM's and LYRA's outputs. Top: scene 5c9274b41b9510f7 (coverage 0.865). Bottom: scene d879362410eaf691 (coverage 0.886).}
    \label{fig:supp_a3}
\end{figure*}

On T\&T, LYRA's matched-protocol result (22.98/0.725/0.174) exceeds PRISM's on the identical $N=8$ clips (21.68/0.645/0.288) across all three metrics. The cost of that quality margin is direct: LYRA takes $\sim$22.4 hours to process these 8 clips against PRISM's $\sim$4.8 minutes, a $277\times$ difference consistent with the per-scene figures in Sec.~3.1 (2.8h vs.\ 36.2s). PRISM's contribution on T\&T is therefore this inference speedup at broadly comparable quality, not superior reconstruction fidelity. 
On the full $N{=}40$ RE10K ablation subset, the comparison narrows further: PRISM (21.28/0.763/0.242) exceeds LYRA (20.97/0.756/0.211) by 0.31~dB on PSNR and is essentially tied on SSIM, while LYRA remains ahead on LPIPS (0.211 vs.\ 0.242), consistent with the perceptual-diversity gap discussed in Sec.~7.
This LPIPS margin still costs LYRA the same $277\times$ time overhead, now $\sim$112~hours against PRISM's $\sim$24~minutes across all 40 scenes. Taken together, PRISM's contribution on both benchmarks is inference speed at comparable-to-superior rendering quality, not a uniform fidelity advantage over LYRA's generative pipeline.

\noindent\textbf{Sensitivity to the frame-sampling interval.} The exact frame-sampling protocol used by LYRA, Wonderland, and Bolt3D~\cite{szymanowicz2025bolt3d} on T\&T is not specified in the corresponding papers, so the interval between the reference frame and each evaluated target frame is a free parameter. To characterize how sensitive PRISM's T\&T result is to this choice, we re-evaluate PRISM with the reference-to-target frame interval widened to every 2nd, 3rd, and 4th frame instead of every consecutive frame, which increases the camera baseline between the reference and each target view.
As Table~\ref{tab:supp_stride} shows, PRISM's T\&T PSNR falls monotonically from 21.98\,dB at interval 1 to 16.84\,dB at interval 4, a 5.14\,dB range driven by the frame-sampling interval alone rather than any change to the model.

\begin{table}[tbp]
\centering
\caption{PRISM on T\&T under an increasing reference-to-target frame interval (wider camera baseline between reference and target views).}
\label{tab:supp_stride}
\begin{tabular}{lccc}
\toprule
Frame interval & PSNR$\uparrow$ & SSIM$\uparrow$ & LPIPS$\downarrow$ \\
\midrule
1 & 21.98 & 0.637 & 0.288 \\
2 & 19.65 & 0.576 & 0.347 \\
3 (Table~1) & 18.29 & 0.560 & 0.369 \\
4 & 16.84 & 0.520 & 0.430 \\
\bottomrule
\end{tabular}
\end{table}

% #---# R2: zoomed PRISM-vs-LYRA crops on real RE10K at low coverage.
% #---# Two scenes from the 0.85-0.90 band, same matched pipeline as above.
\noindent\textbf{Qualitative comparison at low-coverage viewpoints.} To directly visualize the effect of low warp coverage, we select two RE10K scenes with real-view coverage in the 0.85--0.90 range at a frame interval of 3, wider than the stride of 2 used elsewhere on RE10K so that coverage is low enough for the effect to be visible (5c9274b41b9510f7, coverage 0.865, PSNR/SSIM/LPIPS 18.59/0.689/0.400; d879362410eaf691, coverage 0.886, PSNR/SSIM/LPIPS 19.76/0.755/0.451) and render both PRISM and LYRA's matched-protocol pipeline under this setting. Fig.~\ref{fig:supp_a3} shows zoomed crops with disoccluded regions marked in red. In both scenes, PRISM and LYRA show comparable degradation within these regions: softened detail and reduced sharpness relative to ground truth. The visual cost of low coverage is therefore shared across both the feed-forward and generative pipelines, not specific to PRISM's deterministic residual correction.

% #---# Moved from the Conclusion to free main-text space; substance kept.
% #---# Its own section at the end, so no section numbering shifts.
\section{Limitations}

PRISM applies fixed encoder capacity across all scenes, so reconstruction becomes harder as warp coverage drops or scene content grows more complex, as quantified in Sec.~4.
Moreover, the residual encoder is deterministic: in heavily disoccluded regions, it tends to regress toward a conditional-mean completion rather than sampling a sharp, plausible outcome.
Future work could pursue adaptive capacity allocation, such as scaling encoder depth to scene complexity, and lightweight stochastic decoding for low-coverage regions.

\end{document}